\pdfoutput=1

\documentclass[11pt]{article}

\usepackage[preprint]{acl}

\usepackage{xcolor}      %
\usepackage{multirow}    %
\usepackage{booktabs}
\usepackage{arydshln} 
\usepackage{times}
\usepackage{latexsym}
\usepackage{graphicx} %
\usepackage{caption}  %
\usepackage{subcaption} %
\usepackage{xcolor,colortbl}

\definecolor{mygreen1}{RGB}{235,255,235} 
\definecolor{mygreen2}{RGB}{210,255,210} 
\definecolor{mygreen3}{RGB}{185,255,185}

\definecolor{myred1}{RGB}{255,235,235} 
\definecolor{myred2}{RGB}{255,210,210}
\definecolor{myred3}{RGB}{255,185,185}

\usepackage[T1]{fontenc}

\usepackage[utf8]{inputenc}

\usepackage{microtype}
\usepackage{amsmath}
\usepackage{graphicx} %
\usepackage{amsfonts}

\usepackage{inconsolata}

\usepackage{graphicx}

\title{Source-primed Multi-turn Conversation Helps Large Language Models Translate Documents}

\author{
    Hanxu Hu, Jannis Vamvas, Rico Sennrich \\
    University of Zurich \\
    \small\texttt{\{hanxu.hu, jannisnikos.vamvas, rico.sennrich\}@uzh.ch}
}

\begin{document}
\maketitle
\begin{abstract}
LLMs have paved the way for truly simple document-level machine translation, but challenges such as omission errors remain. In this paper, 
we study a simple method for handling document-level machine translation, by leveraging previous contexts in a multi-turn conversational manner.
Specifically, by decomposing documents into segments and iteratively translating them while maintaining previous turns, this method ensures coherent translations without additional training, and can fully re-use the KV cache of previous turns thus minimizing computational overhead. We further propose a `source-primed' method that first provides the whole source document before multi-turn translation. We empirically show this multi-turn method outperforms both translating entire documents in a single turn and translating each segment independently according to multiple automatic metrics in representative LLMs, establishing a strong baseline for document-level translation using LLMs. \footnote{Code and data are available here: \url{https://github.com/ZurichNLP/multiturn-llm-docmt}}

\end{abstract}

\section{Introduction}
Large language models (LLMs) have demonstrated notable abilities in handling various natural language processing tasks and following diverse instructions effectively. One key application is machine translation, whereas previous approaches relied on specialized encoder-decoder translation models. Recent studies have explored both prompting techniques \cite{chen-etal-2024-iterative, karpinska-iyyer-2023-large, lu-etal-2024-chain} and fine-tuning methods \cite{tower, wu2024adaptinglargelanguagemodels} for improving LLMs' translation capabilities.
While document-level translation remains more challenging than sentence-level tasks \cite{kocmi-etal-2024-findings}, recent works have explored ways to enhance LLMs' document translation abilities through fine-tuning on parallel datasets \cite{wu2024adaptinglargelanguagemodels} and various prompting techniques \cite{wang-etal-2023-document-level} to improve context awareness. These approaches provide comprehensive analyses of how current LLMs handle document-level translation.

Multi-turn conversation is a representative way of interacting with LLMs, and it is a natural ability that LLMs gain during the post-training stage. Previous works have leveraged this ability for following instructions \cite{zheng2023judging} and building agents \cite{park2024versatilemotionlanguagemodels} for reasoning tasks. More recently, methods have been explored that leverage multiple segments as exemplars in a multi-turn manner to help language models with translation tasks, but these segments tend to be fixed \cite{kocmi-etal-2024-findings} and do not adapt to subsequent translation data. While retrieval-based approaches have been proposed \cite{zhang23,zafar-etal-2024-setu-adapt}, they do not make use of context from the same document. \cite{wang-etal-2023-document-level} first explored the setting of inputting a document's content sentence-by-sentence in a multi-turn manner. In this paper, we extend this setting to translate documents and use paragraphs as the translation unit in each turn.

\begin{figure*}[t]
    \centering
    \includegraphics[width=\linewidth]{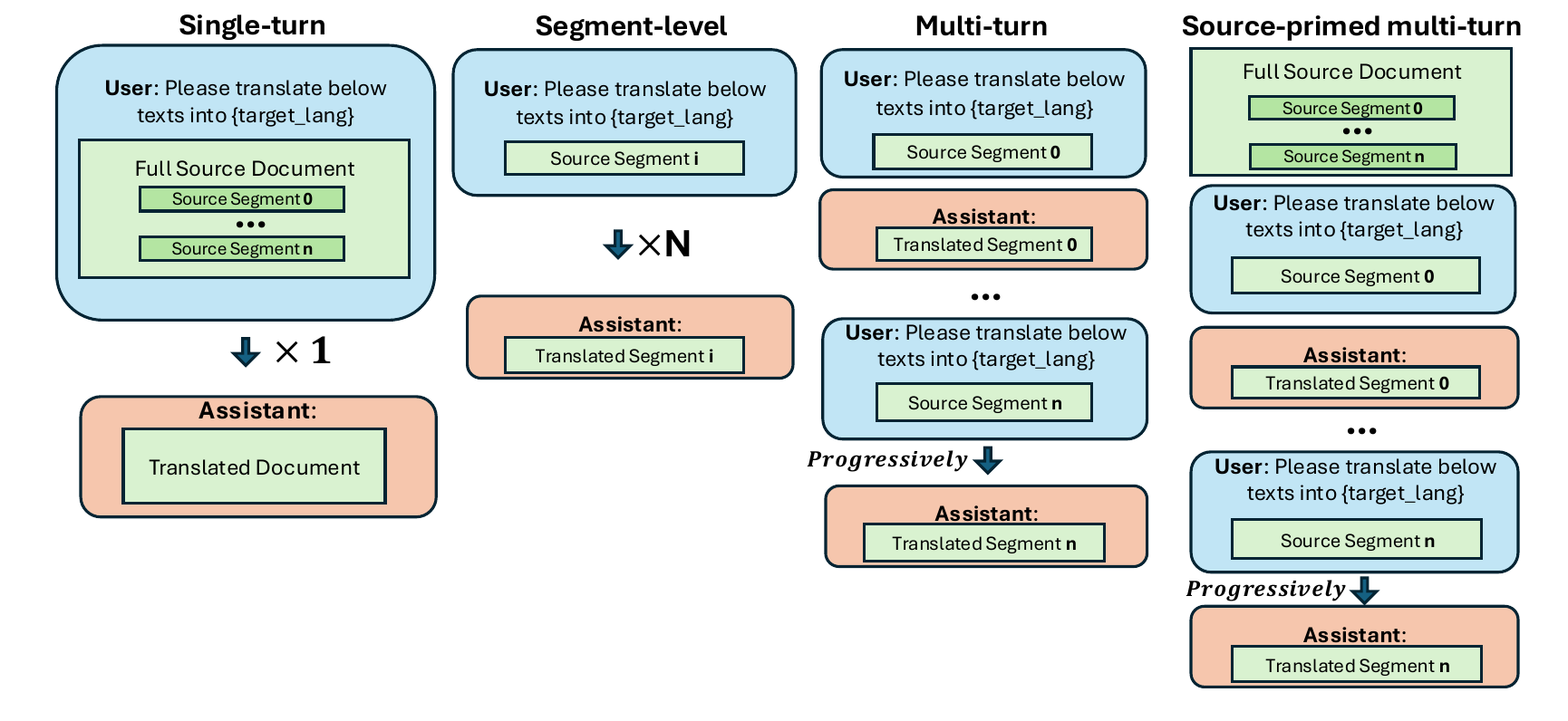}
    \caption{Different settings of document-level translation using LLMs.}
\end{figure*}

Specifically, we separate the document into multiple segments, and then in each conversational turn, we pass a segment combined with user instructions as input, having the model output the corresponding translated segment. LLMs have access to all previous turns, which helps translate the current segment. In this manner,
we merely expand the conversation without modifying the prefix, allowing for caching of previous turns.
This multi-turn strategy has the theoretical drawback compared to single-turn strategies in that initial segments are translated with little context.
Thus, we propose a new variant that first provides the whole source document, then conducts multi-turn paragraph-level translation. This gives the model access to future context that previous multi-turn approaches lack. This approach has the advantage of providing information about the document's topic and style, which might help the model generate appropriate tense and formality levels from the start.

This methodology is simple and effective, does not require additional training, and can be applied to any LLMs that support chat mode. We conducted comprehensive experiments on representative proprietary and open-weight LLMs to show the effectiveness of our method compared to the original multi-turn setting and two other baseline settings: inputting the whole document in a single turn and inputting each segment without context at each inference step. The BLEU, COMET and BlonDE \cite{blonde} scores indicate that the multi-turn conversational method performs better than other settings and can serve as a standard for future document translation research. Furthermore, we conduct a comprehensive analysis of how different factors affect this multi-turn conversational method, such as domain type and document length.

\section{Methodology}
\subsection{Baselines}
\subsubsection{Single-turn document-level translation}
The simplest way of translating a document is to input the whole source document and translate it at once in a single turn. It allows the model to see the context of the entire document, thereby ensuring coherence to a certain extent, but it relies on the model's ability of long-context understanding and generation.
\subsubsection{Segment-level translation }
Another approach of translating a document is to first split it into segments and then translate each segment independently, as is done in the WMT24 shared task \cite{kocmi-etal-2024-findings}. Segment-level translation has been standard for decades, but it lacks global information of document and might have reduced coherence. There has been a recent trend to use paragraphs instead of sentences as segments \cite{karpinska-iyyer-2023-large,kocmi-etal-2024-findings}.
\subsection{Multi-turn conversational translation}
Segment-in-context translation has the potential of allowing the LLM to benefit from document-level context while the segment units hopefully minimize omission errors. 
Following the prompting strategy of \citet{wang-etal-2023-document-level}, we aim for efficient scalability to long documents, and thus frame document translation as a multi-turn conversation where previous turns can be accessed but are never overwritten, thus potentially allowing an LLM to compute each hidden state only once and cache hidden states from previous turns. As an added benefit, LLMs are already trained on such multi-turn conversations, potentially allowing us to use this framework without extra fine-tuning. Additionally, we introduce a variant named as \textbf{`source primed multi-turn'} that first provides the whole source document before conducting multi-turn translation, which gives the model access to future context.

\begin{table*}[h!]
\centering
\begin{tabular}{@{}lcccccc@{}}
\toprule
\textbf{Method} & \multicolumn{2}{c}{\textbf{Qwen-2.5-7B-Instruct}} & \multicolumn{2}{c}{\textbf{Llama-3.1-8B-Instruct}} & \multicolumn{2}{c}{\textbf{GPT-4o-mini}} \\
\cmidrule(lr){2-3} \cmidrule(lr){4-5} \cmidrule(lr){6-7}
                & dBLEU & COMET\_da & dBLEU & COMET\_da & dBLEU & COMET\_da \\
\midrule
Single-turn         & 20.83 & -     & 20.68 & -     & 31.98 & -     \\
Single-turn + ICL   & 20.66 & -     & 21.79 & -     & 31.95 & -     \\
Segment-level       & 18.72 & 70.36 & 21.65 & 76.40 & 31.82 & 83.04 \\
Segment-level + ICL & 20.52 & 71.98 & 23.07 & 78.40 & 32.37 & 84.08 \\
Multi-turn          & 20.37 & 72.37 & 22.85 & 78.40 & 31.74 & 84.30 \\
Multi-turn + ICL    & 21.23 & 72.96 & 23.47 & 79.02 & 32.50 & 
84.30 \\
Multi-turn sp (Ours)  & \textbf{21.46} &  \textbf{74.23} & 23.22 & 78.95 & 32.51 &  84.38 \\
Multi-turn sp + ICL (Ours)  & 21.10  & 73.86 & \textbf{24.23} & \textbf{79.20} & \textbf{32.73}  &\textbf{84.42}  \\
\bottomrule
\end{tabular}
\caption{Results on WMT-24 (Average across all directions), where ICL means adding few-shot in-context learning exemplars as prefix. We do not report COMET in the single-turn setting because we lack segment-level alignments. Multi-turn sp means our source-primed multi-turn variant.}
\label{tab:main}
\end{table*}
\hspace{20mm}

\section{Experiments}

\subsection{Models}
We only used decoder-only instruction models to conduct all experiments, due to their general-purpose nature and design for chat usage, which aligns with our experimental setting. We use the proprietary model GPT-4o-mini \cite{openai2024gpt4ocard}, and the open-weight instruction models Llama-3.1-8B-Instruct \cite{dubey2024llama} and Qwen-2.5-7B-Instruct \cite{qwen2025qwen25technicalreport}. We run the open-weight models using the vLLM framework with greedy search.\footnote{\url{https://github.com/vllm-project/vllm}}
\subsection{Tasks}
We experiment with methods mainly on the WMT 24 General Track for general document-level machine translation scenarios \cite{kocmi-etal-2024-findings}, which is segmented and aligned at the paragraph level. It contains 9 En-to-X (English to Chinese, Czech, German, Hindi, Icelandic, Japanese, Russian, Spanish, Ukrainian) directions and 2 X-to-X (Czech to Ukrainian and Japanese to Chinese) directions. It covers various domains (news, literary, speech, social media) and has documents of varying lengths (different numbers of paragraphs). For further evaluation with the document-level translation metric BlonDE \cite{blonde}, we also use the Chinese-to-English direction from WMT 23 \cite{kocmi-etal-2023-findings} to evaluate our methods and baselines.

\subsection{Evaluation Metrics}
We use the COMET-22 default model \cite{rei-etal-2022-comet} and sacreBLEU \cite{post-2018-call} implementation of BLEU to evaluate translation quality. For COMET, due to its context length limit, we evaluate each segment independently and report average scores; for BLEU, we consider n-gram matches at the document level \cite{liu-etal-2020-multilingual-denoising}. Additionally, we use BlonDE \cite{blonde} to evaluate document-level translation, which specifically measures the correctness of features that are known to benefit from wider context in Chinese-to-English translation, such as tense correctness, pronouns, transliteration, entities, and connectives.

\subsection{Results}
We report our main results in Table \ref{tab:main}. It shows average scores of document-level BLEU and COMET-22 \cite{rei-etal-2022-comet} across all directions in the WMT-24 general track. Following the setting of the WMT-24 shared task \citet{kocmi-etal-2024-findings}, we also report results with in-context learning (ICL) where we provide the same prompt with 3 exemplars as the WMT-24 shared task did. Detailed results for each language direction are in Appendix Table \ref{tab:wmt24-comparison1} and Table \ref{tab:wmt24-comparison2}.

The results show that all four multi-turn conversational methods achieve better translation performance compared to both segment-level methods and single-turn methods across both metrics. The setting of multi-turn with first providing the whole source document (Multi-turn sp) achieves the best results in all cases, demonstrating the advantage of our proposed variant.

Additionally, we report results for the Zh-En direction in WMT-23 using Llama-3.1-8B-Instruct (Table \ref{tab:zh-en}). The document-level metric BlonDE shows a clearer performance gap between multi-turn and segment-level translation compared to dBLEU.

\begin{table}[]
\centering
\small
\begin{tabular}{@{}lcc@{}}
\toprule
\multicolumn{1}{c}{\textbf{Method / Metrics}} & \textbf{dBLEU} & \textbf{BlonDE} \\ \midrule
\multicolumn{3}{c}{\textbf{Llama-3.1-8B-Instruct}} \\ \cmidrule(lr){1-3}
\textbf{Single-turn}                                 & 23.67          & -               \\
\textbf{Single-turn + ICL}                           & 22.47          & -               \\
\textbf{Segment-level}                               & 24.44          & 33.67           \\
\textbf{Segment-level + ICL}                         & 24.92          & 36.61           \\
\textbf{Multi-turn}                                 & 25.04          & 34.35          \\
\textbf{Multi-turn + ICL}                           & 25.49 & 38.03  \\ %
\textbf{Multi-turn sp }                        & 25.42 & 37.91  \\ %
\textbf{Multi-turn sp + ICL}               &   \textbf{25.82}      &   \textbf{38.75}     \\ \bottomrule
\end{tabular}
\caption{Results on WMT-23 Zh-En direction evaluated with dBLEU and BlonDE.}
\label{tab:zh-en}
\end{table}
\noindent

\begin{table}[]
\centering
\small
\begin{tabular}{@{}cll@{}}
\toprule
\textbf{Domains / Settings} & \multicolumn{1}{c}{\textbf{Seg-level ICL}} & \multicolumn{1}{c}{\textbf{Multi-turn ICL}} \\ \midrule
\textbf{Literary}  & 21.52 & \cellcolor{mygreen1}{22.22 (+0.70)} \\
\textbf{News}      & 24.59 & \cellcolor{mygreen2}{25.57 (+0.98)} \\
\textbf{Social}    & 22.07 & \cellcolor{myred2}{20.11 (-1.96)}   \\
\textbf{Speech}    & 24.03 & \cellcolor{mygreen1}{24.25 (+0.22)} \\
\textbf{Personal}  & 20.37 & \cellcolor{myred1}{20.09 (-0.28)}   \\
\textbf{Education} & 26.11 & \cellcolor{mygreen3}{30.27 (+4.16)} \\
\textbf{Voice}     & 22.45 & \cellcolor{mygreen1}{22.70 (+0.25)} \\
\textbf{Official}  & 24.95 & \cellcolor{mygreen1}{25.14 (+0.19)} \\
\bottomrule
\end{tabular}
\caption{Results on WMT-24, showing average dBLEU score across all language directions in different domains. }
\label{tab:domains}
\end{table}
\noindent

\section{Analysis}

\subsection{Results for different domains}
In Table \ref{tab:domains}, we report the results of using Llama-3.1-8B-Instruct on WMT-24 for different domains across all language pairs. We find that in the domains of Literary, Education, and News, using the multi-turn conversational method performs better than single-turn and segment-level translation. The largest improvement (+4.16 dBLEU) is in the educational domain, which appears only in the Czech-Ukrainian direction and consists of exercises from elementary-school students across various subjects, where segments are short and require context for disambiguation.

\subsection{Omission Errors in Long Documents}
\begin{figure}[htbp]
    \begin{subfigure}[t]{0.44\textwidth}
        \centering
        \includegraphics[width=\textwidth]{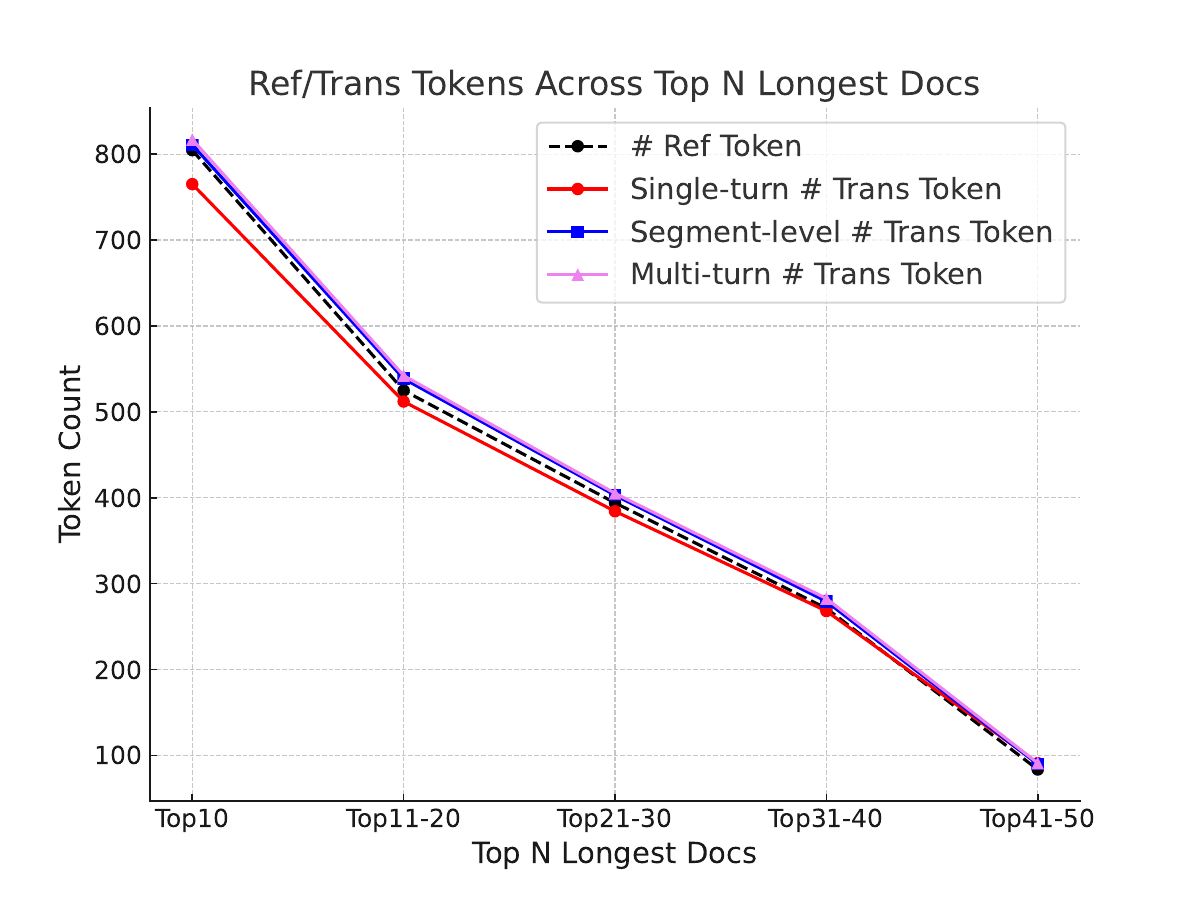}
    \end{subfigure}
    \caption{Token number across top N longest docs}
    \label{fig:topn}
\end{figure}

We visualize the results of different strategies on long documents in Figure \ref{fig:topn}. Specifically, we report the number of reference tokens and translation hypothesis tokens in the top N longest documents. For the longest 10 documents, we observe a clear gap between reference length and translation hypothesis length in single-turn translation, indicating omission errors. Both segment-level and multi-turn translation help mitigate this problem.

\noindent
\vspace{-5mm}

\subsection{Comparison with Open Submissions of WMT-24 General Task}
We evaluate representative open submissions of current state-of-the-art models for WMT-24 general tasks using COMET and dBLEU, and report the results in Appendix Table~\ref{tab:wmt24-submission}. We used the same prompts as \citet{kocmi-etal-2024-findings}. These results primarily demonstrate that we used competitive models, even though we chose smaller models for efficiency reasons.

\section{Related Work}
\subsection{Large Language Models For Translation}
Large language models have become widely used in a series of language tasks including machine translation \cite{zhu-etal-2024-multilingual}. There are various methods to improve the translation performance of LLMs, including prompting, using in-context learning techniques \cite{chen-etal-2024-iterative, lu-etal-2024-chain}, and creating instruction tuning data for machine translation to enhance their capabilities \cite{tower}.
\vspace{-2mm}
\subsection{Document Level and Context Aware Machine Translations with LLMs}
There are works focusing on using LLMs for document-level machine translation \cite{wang-etal-2023-document-level, cui-etal-2024-efficiently, mohammed2024analyzingcontextutilizationllms, wu2024adaptinglargelanguagemodels}. \citet{chapter2chapter} uses LLMs to handle chapter-level translation and construct related datasets. \citet{kudo-etal-2024-document} decodes multiple sentences at each step and selects the most probable one sequentially, which also leverages context information. \citet{luo-etal-2024-context} uses context information to help document translation, but differs from our multi-turn setting in that their prefix exemplars are not fixed, preventing full KV-cache reuse. \citet{wang-etal-2023-document-level} first explores a multi-turn setting similar to ours but focuses on comparing different numbers of sentences in each turn. \citet{karpinska-iyyer-2023-large} compares paragraph-level translation and sentence-level translation for literary translation and finds paragraph-level translation to be superior.
\vspace{-2mm}
\subsection{Chat Translation}
In the WMT24 Chat shared task \cite{mohammed-etal-2024-findings}, multiple submissions explore segment-in-context prompts \cite{pombal-etal-2024-improving}, with a mix of full document context and sliding window approaches \cite{yang-etal-2024-optimising, sung-etal-2024-context}. While the latter only requires O(n) tokens to be processed, we aim for full document-level access with our multi-turn framework.

\section{Conclusion}
Source-primed multi-turn translation: 1) gives structure to the translation task, reducing omission errors; 2) provides access to the full source document context at the beginning compared to original multi-turn translation, enabling LLMs to improve coherence and other document-level aspects of translation; and 3) allows for caching attention keys and values from previous turns, eliminating the efficiency bottleneck of previous segment-in-context prompting strategies.

\section{Limitations} 
In this paper, we only use automatic metrics for evaluating document level machine translation due to lack of resources. Our evaluation includes a document level metric (BlonDE), but human evaluation remains important future work.

Our efficiency argument is a theoretical one based on the ability to re-use prefixes in our multi-turn strategy, compared to other segment-in-context prompts which require re-processing the full prompt. While we did not implement KV caching ourselves, we note that the GPT-4o-mini API currently implements prompt caching.\footnote{\url{https://platform.openai.com/docs/guides/prompt-caching}, accessed on 12.02.2025.}

\section*{Acknowledgments}
This work was funded by the Swiss National Science Foundation (project InvestigaDiff; no.~10000503).

\bibliography{custom}

\begin{thebibliography}{29}
\providecommand{\natexlab}[1]{#1}

\bibitem[{Alves et~al.(2024)Alves, Pombal, Guerreiro, Martins, Alves, Farajian, Peters, Rei, Fernandes, Agrawal, Colombo, de~Souza, and Martins}]{tower}
Duarte~Miguel Alves, Jos{\'e} Pombal, Nuno~M Guerreiro, Pedro~Henrique Martins, Jo{\~a}o Alves, Amin Farajian, Ben Peters, Ricardo Rei, Patrick Fernandes, Sweta Agrawal, Pierre Colombo, Jos{\'e} G.~C. de~Souza, and Andre Martins. 2024.
\newblock \href {https://openreview.net/forum?id=EHPns3hVkj} {Tower: An open multilingual large language model for translation-related tasks}.
\newblock In \emph{First Conference on Language Modeling}.

\bibitem[{Chen et~al.(2024)Chen, Guo, Haddow, and Heafield}]{chen-etal-2024-iterative}
Pinzhen Chen, Zhicheng Guo, Barry Haddow, and Kenneth Heafield. 2024.
\newblock \href {https://aclanthology.org/2024.eamt-1.17} {Iterative translation refinement with large language models}.
\newblock In \emph{Proceedings of the 25th Annual Conference of the European Association for Machine Translation (Volume 1)}, pages 181--190, Sheffield, UK. European Association for Machine Translation (EAMT).

\bibitem[{Cui et~al.(2024)Cui, Du, Zhu, and Xiong}]{cui-etal-2024-efficiently}
Menglong Cui, Jiangcun Du, Shaolin Zhu, and Deyi Xiong. 2024.
\newblock \href {https://doi.org/10.18653/v1/2024.findings-acl.646} {Efficiently exploring large language models for document-level machine translation with in-context learning}.
\newblock In \emph{Findings of the Association for Computational Linguistics: ACL 2024}, pages 10885--10897, Bangkok, Thailand. Association for Computational Linguistics.

\bibitem[{Dubey et~al.(2024)Dubey, Jauhri, Pandey, Kadian, Al-Dahle, Letman, Mathur, Schelten, Yang, Fan et~al.}]{dubey2024llama}
Abhimanyu Dubey, Abhinav Jauhri, Abhinav Pandey, Abhishek Kadian, Ahmad Al-Dahle, Aiesha Letman, Akhil Mathur, Alan Schelten, Amy Yang, Angela Fan, et~al. 2024.
\newblock The llama 3 herd of models.
\newblock \emph{arXiv preprint arXiv:2407.21783}.

\bibitem[{Jiang et~al.(2022)Jiang, Liu, Ma, Zhang, Yang, Huang, Sennrich, Cotterell, Sachan, and Zhou}]{blonde}
Yuchen Jiang, Tianyu Liu, Shuming Ma, Dongdong Zhang, Jian Yang, Haoyang Huang, Rico Sennrich, Ryan Cotterell, Mrinmaya Sachan, and Ming Zhou. 2022.
\newblock \href {https://doi.org/10.18653/v1/2022.naacl-main.111} {{BlonDe}: An automatic evaluation metric for document-level machine translation}.
\newblock In \emph{Proceedings of the 2022 Conference of the North American Chapter of the Association for Computational Linguistics: Human Language Technologies}, pages 1550--1565, Seattle, United States. Association for Computational Linguistics.

\bibitem[{Jin et~al.(2024)Jin, An, and Ma}]{chapter2chapter}
Linghao Jin, Li~An, and Xuezhe Ma. 2024.
\newblock \href {https://arxiv.org/abs/2407.08978} {Towards chapter-to-chapter context-aware literary translation via large language models}.
\newblock \emph{Preprint}, arXiv:2407.08978.

\bibitem[{Karpinska and Iyyer(2023)}]{karpinska-iyyer-2023-large}
Marzena Karpinska and Mohit Iyyer. 2023.
\newblock \href {https://doi.org/10.18653/v1/2023.wmt-1.41} {Large language models effectively leverage document-level context for literary translation, but critical errors persist}.
\newblock In \emph{Proceedings of the Eighth Conference on Machine Translation}, pages 419--451, Singapore. Association for Computational Linguistics.

\bibitem[{Kocmi et~al.(2024)Kocmi, Avramidis, Bawden, Bojar, Dvorkovich, Federmann, Fishel, Freitag, Gowda, Grundkiewicz, Haddow, Karpinska, Koehn, Marie, Monz, Murray, Nagata, Popel, Popovi{\'c}, Shmatova, Steingr{\'\i}msson, and Zouhar}]{kocmi-etal-2024-findings}
Tom Kocmi, Eleftherios Avramidis, Rachel Bawden, Ond{\v{r}}ej Bojar, Anton Dvorkovich, Christian Federmann, Mark Fishel, Markus Freitag, Thamme Gowda, Roman Grundkiewicz, Barry Haddow, Marzena Karpinska, Philipp Koehn, Benjamin Marie, Christof Monz, Kenton Murray, Masaaki Nagata, Martin Popel, Maja Popovi{\'c}, Mariya Shmatova, Steinth{\'o}r Steingr{\'\i}msson, and Vil{\'e}m Zouhar. 2024.
\newblock \href {https://doi.org/10.18653/v1/2024.wmt-1.1} {Findings of the {WMT}24 general machine translation shared task: The {LLM} era is here but {MT} is not solved yet}.
\newblock In \emph{Proceedings of the Ninth Conference on Machine Translation}, pages 1--46, Miami, Florida, USA. Association for Computational Linguistics.

\bibitem[{Kocmi et~al.(2023)Kocmi, Avramidis, Bawden, Bojar, Dvorkovich, Federmann, Fishel, Freitag, Gowda, Grundkiewicz, Haddow, Koehn, Marie, Monz, Morishita, Murray, Nagata, Nakazawa, Popel, Popovi{\'c}, and Shmatova}]{kocmi-etal-2023-findings}
Tom Kocmi, Eleftherios Avramidis, Rachel Bawden, Ond{\v{r}}ej Bojar, Anton Dvorkovich, Christian Federmann, Mark Fishel, Markus Freitag, Thamme Gowda, Roman Grundkiewicz, Barry Haddow, Philipp Koehn, Benjamin Marie, Christof Monz, Makoto Morishita, Kenton Murray, Makoto Nagata, Toshiaki Nakazawa, Martin Popel, Maja Popovi{\'c}, and Mariya Shmatova. 2023.
\newblock \href {https://doi.org/10.18653/v1/2023.wmt-1.1} {Findings of the 2023 conference on machine translation ({WMT}23): {LLM}s are here but not quite there yet}.
\newblock In \emph{Proceedings of the Eighth Conference on Machine Translation}, pages 1--42, Singapore. Association for Computational Linguistics.

\bibitem[{Kudo et~al.(2024)Kudo, Deguchi, Morishita, Fujii, Ito, Ozaki, Natsumi, Sato, Yano, Takahashi, Kimura, Hara, Sakai, and Suzuki}]{kudo-etal-2024-document}
Keito Kudo, Hiroyuki Deguchi, Makoto Morishita, Ryo Fujii, Takumi Ito, Shintaro Ozaki, Koki Natsumi, Kai Sato, Kazuki Yano, Ryosuke Takahashi, Subaru Kimura, Tomomasa Hara, Yusuke Sakai, and Jun Suzuki. 2024.
\newblock \href {https://doi.org/10.18653/v1/2024.wmt-1.14} {Document-level translation with {LLM} reranking: Team-{J} at {WMT} 2024 general translation task}.
\newblock In \emph{Proceedings of the Ninth Conference on Machine Translation}, pages 210--226, Miami, Florida, USA. Association for Computational Linguistics.

\bibitem[{Liu et~al.(2020)Liu, Gu, Goyal, Li, Edunov, Ghazvininejad, Lewis, and Zettlemoyer}]{liu-etal-2020-multilingual-denoising}
Yinhan Liu, Jiatao Gu, Naman Goyal, Xian Li, Sergey Edunov, Marjan Ghazvininejad, Mike Lewis, and Luke Zettlemoyer. 2020.
\newblock \href {https://doi.org/10.1162/tacl_a_00343} {Multilingual denoising pre-training for neural machine translation}.
\newblock \emph{Transactions of the Association for Computational Linguistics}, 8:726--742.

\bibitem[{Lu et~al.(2024)Lu, Yang, Huang, Zhang, Lam, and Wei}]{lu-etal-2024-chain}
Hongyuan Lu, Haoran Yang, Haoyang Huang, Dongdong Zhang, Wai Lam, and Furu Wei. 2024.
\newblock \href {https://doi.org/10.18653/v1/2024.emnlp-main.55} {Chain-of-dictionary prompting elicits translation in large language models}.
\newblock In \emph{Proceedings of the 2024 Conference on Empirical Methods in Natural Language Processing}, pages 958--976, Miami, Florida, USA. Association for Computational Linguistics.

\bibitem[{Luo et~al.(2024)Luo, Guo, Wei, Shang, Li, Wu, Rao, Li, Yang, and Yang}]{luo-etal-2024-context}
Yuanchang Luo, Jiaxin Guo, Daimeng Wei, Hengchao Shang, Zongyao Li, Zhanglin Wu, Zhiqiang Rao, Shaojun Li, Jinlong Yang, and Hao Yang. 2024.
\newblock \href {https://doi.org/10.18653/v1/2024.wmt-1.97} {Context-aware and style-related incremental decoding framework for discourse-level literary translation}.
\newblock In \emph{Proceedings of the Ninth Conference on Machine Translation}, pages 973--979, Miami, Florida, USA. Association for Computational Linguistics.

\bibitem[{Mohammed et~al.(2024)Mohammed, Agrawal, Farajian, Cabarr{\~a}o, Eikema, Farinha, and C.~De~Souza}]{mohammed-etal-2024-findings}
Wafaa Mohammed, Sweta Agrawal, Amin Farajian, Vera Cabarr{\~a}o, Bryan Eikema, Ana~C Farinha, and Jos{\'e}~G. C.~De~Souza. 2024.
\newblock \href {https://doi.org/10.18653/v1/2024.wmt-1.59} {Findings of the {WMT} 2024 shared task on chat translation}.
\newblock In \emph{Proceedings of the Ninth Conference on Machine Translation}, pages 701--714, Miami, Florida, USA. Association for Computational Linguistics.

\bibitem[{Mohammed and Niculae(2024)}]{mohammed2024analyzingcontextutilizationllms}
Wafaa Mohammed and Vlad Niculae. 2024.
\newblock \href {https://arxiv.org/abs/2410.14391} {Analyzing context utilization of llms in document-level translation}.
\newblock \emph{Preprint}, arXiv:2410.14391.

\bibitem[{OpenAI(2024)}]{openai2024gpt4ocard}
OpenAI. 2024.
\newblock \href {https://arxiv.org/abs/2410.21276} {Gpt-4o system card}.
\newblock \emph{Preprint}, arXiv:2410.21276.

\bibitem[{Park et~al.(2024)Park, Choi, and Yun}]{park2024versatilemotionlanguagemodels}
Jeongeun Park, Sungjoon Choi, and Sangdoo Yun. 2024.
\newblock \href {https://arxiv.org/abs/2410.05628} {Versatile motion language models for multi-turn interactive agents}.
\newblock \emph{Preprint}, arXiv:2410.05628.

\bibitem[{Pombal et~al.(2024)Pombal, Agrawal, and Martins}]{pombal-etal-2024-improving}
Jose Pombal, Sweta Agrawal, and Andr{\'e} Martins. 2024.
\newblock \href {https://doi.org/10.18653/v1/2024.wmt-1.100} {Improving context usage for translating bilingual customer support chat with large language models}.
\newblock In \emph{Proceedings of the Ninth Conference on Machine Translation}, pages 993--1003, Miami, Florida, USA. Association for Computational Linguistics.

\bibitem[{Post(2018)}]{post-2018-call}
Matt Post. 2018.
\newblock \href {https://doi.org/10.18653/v1/W18-6319} {A call for clarity in reporting {BLEU} scores}.
\newblock In \emph{Proceedings of the Third Conference on Machine Translation: Research Papers}, pages 186--191, Brussels, Belgium. Association for Computational Linguistics.

\bibitem[{Rei et~al.(2022)Rei, C.~de Souza, Alves, Zerva, Farinha, Glushkova, Lavie, Coheur, and Martins}]{rei-etal-2022-comet}
Ricardo Rei, Jos{\'e}~G. C.~de Souza, Duarte Alves, Chrysoula Zerva, Ana~C Farinha, Taisiya Glushkova, Alon Lavie, Luisa Coheur, and Andr{\'e} F.~T. Martins. 2022.
\newblock \href {https://aclanthology.org/2022.wmt-1.52} {{COMET}-22: Unbabel-{IST} 2022 submission for the metrics shared task}.
\newblock In \emph{Proceedings of the Seventh Conference on Machine Translation (WMT)}, pages 578--585, Abu Dhabi, United Arab Emirates (Hybrid). Association for Computational Linguistics.

\bibitem[{Sung et~al.(2024)Sung, Lee, Kim, and Kim}]{sung-etal-2024-context}
Mingi Sung, Seungmin Lee, Jiwon Kim, and Sejoon Kim. 2024.
\newblock \href {https://doi.org/10.18653/v1/2024.wmt-1.102} {Context-aware {LLM} translation system using conversation summarization and dialogue history}.
\newblock In \emph{Proceedings of the Ninth Conference on Machine Translation}, pages 1011--1015, Miami, Florida, USA. Association for Computational Linguistics.

\bibitem[{Wang et~al.(2023)Wang, Lyu, Ji, Zhang, Yu, Shi, and Tu}]{wang-etal-2023-document-level}
Longyue Wang, Chenyang Lyu, Tianbo Ji, Zhirui Zhang, Dian Yu, Shuming Shi, and Zhaopeng Tu. 2023.
\newblock \href {https://doi.org/10.18653/v1/2023.emnlp-main.1036} {Document-level machine translation with large language models}.
\newblock In \emph{Proceedings of the 2023 Conference on Empirical Methods in Natural Language Processing}, pages 16646--16661, Singapore. Association for Computational Linguistics.

\bibitem[{Wu et~al.(2024)Wu, Vu, Qu, Foster, and Haffari}]{wu2024adaptinglargelanguagemodels}
Minghao Wu, Thuy-Trang Vu, Lizhen Qu, George Foster, and Gholamreza Haffari. 2024.
\newblock \href {https://arxiv.org/abs/2401.06468} {Adapting large language models for document-level machine translation}.
\newblock \emph{Preprint}, arXiv:2401.06468.

\bibitem[{Yang et~al.(2025)Yang, Yang, Zhang, Hui, Zheng, Yu, Li, Liu, Huang, Wei, Lin, Yang, Tu, Zhang, Yang, Yang, Zhou, Lin, Dang, Lu, Bao, Yang, Yu, Li, Xue, Zhang, Zhu, Men, Lin, Li, Tang, Xia, Ren, Ren, Fan, Su, Zhang, Wan, Liu, Cui, Zhang, and Qiu}]{qwen2025qwen25technicalreport}
An~Yang, Baosong Yang, Beichen Zhang, Binyuan Hui, Bo~Zheng, Bowen Yu, Chengyuan Li, Dayiheng Liu, Fei Huang, Haoran Wei, Huan Lin, Jian Yang, Jianhong Tu, Jianwei Zhang, Jianxin Yang, Jiaxi Yang, Jingren Zhou, Junyang Lin, Kai Dang, Keming Lu, Keqin Bao, Kexin Yang, Le~Yu, Mei Li, Mingfeng Xue, Pei Zhang, Qin Zhu, Rui Men, Runji Lin, Tianhao Li, Tianyi Tang, Tingyu Xia, Xingzhang Ren, Xuancheng Ren, Yang Fan, Yang Su, Yichang Zhang, Yu~Wan, Yuqiong Liu, Zeyu Cui, Zhenru Zhang, and Zihan Qiu. 2025.
\newblock \href {https://arxiv.org/abs/2412.15115} {Qwen2.5 technical report}.
\newblock \emph{Preprint}, arXiv:2412.15115.

\bibitem[{Yang et~al.(2024)Yang, Mu, Bontcheva, and Song}]{yang-etal-2024-optimising}
Xinye Yang, Yida Mu, Kalina Bontcheva, and Xingyi Song. 2024.
\newblock \href {https://doi.org/10.18653/v1/2024.wmt-1.101} {Optimising {LLM}-driven machine translation with context-aware sliding windows}.
\newblock In \emph{Proceedings of the Ninth Conference on Machine Translation}, pages 1004--1010, Miami, Florida, USA. Association for Computational Linguistics.

\bibitem[{Zafar et~al.(2024)Zafar, Castaldo, Nayak, Haque, and Way}]{zafar-etal-2024-setu-adapt}
Maria Zafar, Antonio Castaldo, Prashanth Nayak, Rejwanul Haque, and Andy Way. 2024.
\newblock \href {https://doi.org/10.18653/v1/2024.wmt-1.104} {The {SETU}-{ADAPT} submissions to {WMT} 2024 chat translation tasks}.
\newblock In \emph{Proceedings of the Ninth Conference on Machine Translation}, pages 1023--1030, Miami, Florida, USA. Association for Computational Linguistics.

\bibitem[{Zhang et~al.(2023)Zhang, Haddow, and Birch}]{zhang23}
Biao Zhang, Barry Haddow, and Alexandra Birch. 2023.
\newblock Prompting large language model for machine translation: a case study.
\newblock In \emph{Proceedings of the 40th International Conference on Machine Learning}, ICML'23. JMLR.org.

\bibitem[{Zheng et~al.(2023)Zheng, Chiang, Sheng, Zhuang, Wu, Zhuang, Lin, Li, Li, Xing, Zhang, Gonzalez, and Stoica}]{zheng2023judging}
Lianmin Zheng, Wei-Lin Chiang, Ying Sheng, Siyuan Zhuang, Zhanghao Wu, Yonghao Zhuang, Zi~Lin, Zhuohan Li, Dacheng Li, Eric Xing, Hao Zhang, Joseph~E. Gonzalez, and Ion Stoica. 2023.
\newblock \href {https://openreview.net/forum?id=uccHPGDlao} {Judging {LLM}-as-a-judge with {MT}-bench and chatbot arena}.
\newblock In \emph{Thirty-seventh Conference on Neural Information Processing Systems Datasets and Benchmarks Track}.

\bibitem[{Zhu et~al.(2024)Zhu, Liu, Dong, Xu, Huang, Kong, Chen, and Li}]{zhu-etal-2024-multilingual}
Wenhao Zhu, Hongyi Liu, Qingxiu Dong, Jingjing Xu, Shujian Huang, Lingpeng Kong, Jiajun Chen, and Lei Li. 2024.
\newblock \href {https://doi.org/10.18653/v1/2024.findings-naacl.176} {Multilingual machine translation with large language models: Empirical results and analysis}.
\newblock In \emph{Findings of the Association for Computational Linguistics: NAACL 2024}, pages 2765--2781, Mexico City, Mexico. Association for Computational Linguistics.

\end{thebibliography}

\appendix
\newpage
\section{Appendix}
\subsection{Comparison With Submissions of WMT-24 shared task}
\label{sec:appendix}

Below are the results of submissions from representative systems on WMT-24 general tasks and the results of multi-turn and segment-level settings in our experiments, evaluated using dBLEU (document-level) and COMET (paragraph-level).

\begin{table}[h]
\begin{tabular}{@{}lcc@{}}
\toprule
\multicolumn{1}{l}{\textbf{Systems / Metrics}} & \multicolumn{1}{l}{\textbf{dBLEU}} & \multicolumn{1}{l}{\textbf{COMET}} \\ \midrule
\textbf{Claude-3.5}                            & \textbf{33.34}                     & 85.01                              \\
\textbf{GPT-4o-mini (sp MTurn ICL)}                 & 32.73                              & 84.42                             \\
\textbf{GPT-4o-mini (Seg ICL )}                  & 32.37                              & 84.08                              \\
\textbf{Gemini-1.5-Pro}                        & 30.76                              & 83.67                              \\
\textbf{GPT-4}                                 & 30.43                              & 84.05                              \\
\textbf{CommandR-plus}                         & 29.51                              & 82.50                              \\
\textbf{Unbabel-Tower70B}                      & 28.81                              & \textbf{86.46}                              \\
\textbf{Aya23}                                 & 28.07                              & 79.85                              \\
\textbf{Llama3-70B}                            & 27.30                              & 81.32                              \\
\textbf{Mistral-Large}                         & 26.78                              & 80.65                              \\ \bottomrule
\end{tabular}
\caption{Results of representative systems submissions of WMT24.}
\label{tab:wmt24-submission}
\end{table}

\subsection{Discussion of Non-General LLMs}
We also conducted experiments using the machine translation enhanced LLM Tower-Instruct-7B. We report the detailed results of this model in Table \ref{tab:wmt24-comparison1} and Table \ref{tab:wmt24-comparison2}. The results indicate that using the multi-turn method with Tower-Instruct-7B is not effective compared to using segment-level translation, which might be due to TowerInstruct having been mostly optimized on single-turn machine translation data (75\% zero-shot data).
\subsection{Dataset License}
In this paper, we use the WMT-24 and WMT-23 test sets, which comply with their respective license rules and can be freely used for research purposes.

\begin{table*}[ht]
    \centering
    \small
    \setlength{\tabcolsep}{4pt} %
    \renewcommand{\arraystretch}{1.2} %
    \begin{tabular}{@{}llccccccccccc@{}}
        \toprule
        \textbf{Model} & \textbf{Setup} & \textbf{ja-zh} & \textbf{cs-uk} & \textbf{en-de} & \textbf{en-zh} & \textbf{en-es} & \textbf{en-hi} & \textbf{en-ru} & \textbf{en-is} & \textbf{en-ja} & \textbf{en-cs} & \textbf{en-uk} \\
        \midrule
        \multirow{6}{*}{\textbf{TowerInstruct-7B-v0.2}}
        & Single-turn        & 12.52 &  4.15 & 27.90 & 28.39 & 34.59 &  1.54 & 19.45 &  1.37 &  5.56 &  8.89 &  8.47 \\
        & Single-turn + ICL    & 12.98 & 10.90 & 24.27 & 25.45 & 30.11 &  1.71 & 17.57 &  0.89 &  5.35 &  7.96 & 10.93 \\
        & Segment-level           & 21.87 &  7.28 & 30.85 & 37.01 & 39.94 &  4.87 & 22.46 &  4.53 & 10.22 & 14.30 &  9.26 \\
        & Segment-level + ICL       & 21.79 & 11.40 & 30.37 & 37.08 & 40.71 &  5.54 & 21.62 &  4.50 & 14.80 & 13.23 & 13.03 \\
        & Mturn-turn        & 22.03 &  8.13 & 29.96 & 36.34 & 40.04 &  3.15 & 21.17 &  2.69 &  7.20 & 13.06 &  7.47 \\
        & Mturn-turn + ICL     & 22.36 &  9.69 & 27.67 & 35.90 & 37.18 &  4.03 & 21.04 &  3.04 & 13.31 & 12.88 & 12.37 \\
        \midrule
        \multirow{6}{*}{\textbf{Llama-3.1-8B-Instruct}}
        & Single-turn        & 11.94 & 23.23 & 27.79 & 31.06 & 38.90 & 18.37 & 17.41 &  6.09 & 15.89 & 19.99 & 16.84 \\
        & Single-turn + ICL    & 20.06 & 23.33 & 28.14 & 32.15 & 39.12 & 17.63 & 20.20 &  6.38 & 14.58 & 19.89 & 18.16 \\
        & Segment-level           & 11.47 & 21.30 & 27.24 & 33.26 & 39.12 & 19.59 & 19.37 &  6.80 & 21.53 & 20.64 & 17.89 \\
        & Segment-level + ICL       & 23.02 & 23.02 & 27.79 & 36.80 & 39.97 & 20.03 & 19.65 &  8.54 & 15.76 & 20.80 & 18.40 \\
        & Multi-turn        & 17.32 & 23.51 & 27.94 & 34.87 & 39.69 & 20.50 & 19.70 &  7.85 & 21.90 & 19.83 & 18.22 \\
        & Multi-turn + ICL     & 23.34 & 23.26 & 28.08 & 37.18 & 40.39 & 20.48 & 19.95 &  6.68 & 20.04 & 20.02 & 18.70 \\
        & Multi-turn sp         & 22.78 & 24.06 & 26.72 & 32.21 & 39.75 & 20.63 & 20.21 & 8.72  & 21.17 & 20.02 & 19.18 \\
       & Multi-turn sp + ICL    & 26.49 & 24.10 & 26.94 & 37.41 & 40.13 & 20.96 & 19.74 & 8.93  & 22.31 & 20.80 & 18.67 \\
        \midrule
        \multirow{6}{*}{\textbf{Qwen-2.5-7B-Instruct}}
     & Single-turn                   & 30.10 & 14.90 & 25.80 & 41.02 & 37.98 & 7.71  & 16.69 & 4.53  & 21.46 & 15.57 & 13.33 \\
     & Single-turn + ICL              & 29.24 & 16.46 & 25.56 & 41.22 & 36.96 & 8.12  & 16.24 & 4.11  & 20.50 & 15.37 & 13.51 \\
     & Segment-level                     & 27.85 & 9.58  & 21.98 & 36.65 & 33.43 & 10.34 & 15.39 & 6.43  & 16.80 & 14.49 & 13.01 \\
     & Segment-level + ICL                 & 30.06 & 15.07 & 23.54 & 38.10 & 35.45 & 10.42 & 16.21 & 7.01  & 20.91 & 15.10 & 13.82 \\
     & Multi-turn                    & 28.37 & 13.83 & 24.46 & 37.92 & 36.75 & 10.81 & 16.25 & 7.14  & 18.55 & 16.07 & 13.91 \\
    & Multi-turn + ICL             & 30.50 & 16.30 & 24.43 & 39.57 & 37.16 & 10.69 & 16.69 & 7.08  & 20.47 & 16.16 & 14.48 \\
        & Multi-turn sp         & 30.31 & 16.81 & 25.91 & 41.66 & 37.84 & 10.83 & 14.91 & 4.25  & 22.61 & 17.28 & 13.62 \\
         & Multi-turn sp + ICL    & 30.51 & 16.15 & 24.12 & 41.43 & 37.65 & 10.95 & 14.53 & 4.64  & 22.21 & 17.27 & 12.68 \\
        \midrule
        \multirow{6}{*}{\textbf{GPT-4o-mini}}
        & Single-turn        & 33.99 & 32.48 & 34.10 & 45.11 & 45.75 & 26.79 & 23.85 & 20.91 & 29.55 & 29.94 & 29.28 \\
        & Single-turn + ICL    & 33.65 & 32.19 & 35.09 & 45.14 & 45.80 & 26.93 & 22.78 & 21.28 & 29.04 & 30.10 & 29.44 \\
        & Segment-level           & 33.52 & 33.17 & 34.31 & 44.35 & 45.65 & 26.64 & 23.59 & 21.12 & 29.17 & 29.05 & 29.48 \\
        & Segment-level + ICL       & 34.04 & 32.89 & 34.87 & 44.51 & 46.29 & 27.42 & 24.16 & 22.04 & 29.91 & 29.85 & 30.11 \\
        & Multi-turn         & 34.51 & 33.31 & 34.60 & 44.59 & 46.39 & 19.68 & 24.25 & 22.04 & 29.98 & 29.98 & 29.82 \\
        & Multi-turn + ICL     & 34.45 & 33.23 & 34.82 & 44.86 & 46.28 & 27.30 & 24.16 & 22.18 & 30.07 & 29.96 & 30.09 \\
        & Multi-turn sp          & 34.60 & 33.41 & 34.71 & 45.06 & 45.84 & 27.19 & 24.65 & 21.94 & 29.93 & 30.13 & 30.20 \\
        & Multi-turn sp + ICL      & 34.49 & 33.37 & 34.94 & 44.88 & 45.93 & 27.51 & 25.03 & 22.36 & 30.40 & 30.74 & 30.41 \\
        \bottomrule
    \end{tabular}
    \caption{Comparison of dBLEU results for WMT24 across different setups and language pairs for TowerInstruct-7B-v0.2, Llama-3.1-8B-Instruct, Qwen-2.5-7B-Instruct and GPT-4o-mini.}
    \label{tab:wmt24-comparison1}
\end{table*}

\begin{table*}[ht]
    \centering
    \small
    \setlength{\tabcolsep}{4pt} %
    \renewcommand{\arraystretch}{1.2} %
    \begin{tabular}{@{}llccccccccccc@{}}
        \toprule
        \textbf{Model} & \textbf{Setup} 
                       & \textbf{ja-zh} & \textbf{cs-uk} & \textbf{en-de} & \textbf{en-zh} 
                       & \textbf{en-es} & \textbf{en-hi} & \textbf{en-ru} & \textbf{en-cs} 
                       & \textbf{en-is} & \textbf{en-ja} & \textbf{en-uk} \\
        \midrule
        \multirow{4}{*}{\textbf{TowerInstruct-7B-v0.2}}
        & Segment-level      
            & 67.04 & 81.11 & 82.67 & 48.04 & 48.46 & 74.42 & 80.84 & 77.74 & 82.67 & 75.70 & 77.62 \\
        & Segment-level + ICL
            & 66.02 & 80.77 & 82.36 & 46.55 & 43.09 & 76.30 & 80.32 & 72.57 & 82.53 & 75.54 & 78.07 \\
        & Multi-turn           
            & 64.76 & 80.38 & 82.19 & 45.17 & 42.19 & 70.72 & 78.93 & 73.44 & 81.43 & 75.90 & 76.79 \\
        & Multi-turn + ICL     
            & 64.01 & 75.43 & 77.51 & 42.45 & 37.80 & 73.11 & 79.61 & 68.00 & 81.29 & 73.52 & 76.19 \\
            \midrule
        \multirow{4}{*}{\textbf{Llama-3.1-8B-Instruct}} 
        & Segment-level        
            & 75.82 & 80.73 & 78.10 & 79.41 & 80.31 & 71.39 & 77.36 & 78.45 & 59.48 & 81.66 & 77.71 \\
        & Segment-level + ICL  
            & 79.70 & 82.08 & 78.95 & 82.39 & 81.56 & 73.04 & 79.28 & 78.93 & 62.68 & 83.51 & 80.25 \\
        & Multi-turn           
            & 79.84 & 83.32 & 79.58 & 82.03 & 81.52 & 72.94 & 78.64 & 79.82 & 61.92 & 83.61 & 79.12 \\
        & Multi-turn + ICL     
            & 81.55 & 83.70 & 79.70 & 82.94 & 82.23 & 73.79 & 79.46 & 79.90 & 62.10 & 84.26 & 79.61 \\
    &Multi-turn sp       & 80.84 & 84.32 & 80.07 & 82.84 & 82.05 & 73.47 & 79.77 & 80.46 & 61.79 & 83.55 & 79.28 \\
    &Multi-turn sp + ICL & 81.42 & 84.28 & 80.26 & 83.03 & 82.31 & 73.55 & 79.69 & 80.51 & 62.79 & 84.27 & 80.13 \\
        \midrule
        \multirow{4}{*}{\textbf{Qwen-2.5-7B-Instruct}} 
&Segment-level            & 80.84 & 68.21 & 75.58 & 82.25 & 77.79 & 55.79 & 73.66 & 70.65 & 45.58 & 76.83 & 66.76 \\
&Segment-level + ICL      & 82.39 & 72.39 & 76.53 & 82.99 & 79.26 & 58.08 & 73.52 & 71.36 & 46.91 & 80.41 & 67.89 \\
&Multi-turn               & 81.93 & 72.15 & 77.75 & 83.57 & 80.39 & 58.84 & 74.27 & 72.93 & 47.55 & 78.44 & 68.22 \\
&Multi-turn + ICL         & 83.07 & 73.69 & 77.89 & 83.82 & 80.61 & 58.99 & 73.61 & 72.89 & 47.77 & 80.63 & 69.61 \\
&Multi-turn sp       & 83.00 & 75.07 & 78.96 & 84.32 & 81.42 & 61.35 & 76.45 & 74.78 & 48.25 & 82.59 & 70.33 \\
&Multi-turn sp + ICL & 83.16 & 74.30 & 78.64 & 84.10 & 81.18 & 61.42 & 75.74 & 74.15 & 48.10 & 81.85 & 69.82 \\
        \midrule
        \multirow{4}{*}{\textbf{GPT-4o-mini}}
        & Segment-level        
            & 83.60 & 88.83 & 82.59 & 84.58 & 82.99 & 76.78 & 81.65 & 84.18 & 78.10 & 86.00 & 84.13 \\
        & Segment-level + ICL  
            & 84.29 & 89.00 & 83.00 & 84.84 & 84.33 & 78.49 & 83.00 & 85.78 & 79.46 & 86.92 & 85.73 \\
        & Multi-turn          
            & 84.74 & 89.24 & 83.11 & 85.23 & 84.59 & 78.44 & 83.36 & 86.05 & 79.63 & 87.20 & 85.75 \\
        & Multi-turn + ICL     
            & 84.77 & 89.23 & 83.14 & 85.32 & 84.57 & 78.46 & 83.23 & 86.03 & 79.54 & 87.22 & 85.78 \\
            & Multi-turn sp       & 84.80 & 89.24 & 83.06 & 85.35 & 84.52 & 78.40 & 83.38 & 86.10 & 79.23 & 87.51 & 85.63 \\
            & Multi-turn sp + ICL & 84.73 & 89.26 & 83.24 & 85.36 & 84.45 & 78.50 & 83.47 & 86.32 & 79.79 & 87.72 & 85.80 \\        \bottomrule
    \end{tabular}
    \caption{Comparison of COMET score for WMT-24 across different setups and language pairs for Tower-7B-Instruct, Llama-3.1-8B-Instruct, Qwen-2.5-7B-Instruct, and GPT-4o-mini.}
    \label{tab:wmt24-comparison2}
\end{table*}

\end{document}